\documentclass[sigconf]{acmart} 
\usepackage{graphicx}
\usepackage{amsmath}
\usepackage{subcaption}
\usepackage{epstopdf}
\usepackage{amsthm}
\usepackage{booktabs}
\usepackage{hyperref}
\usepackage{multirow}
\usepackage{romannum}
\usepackage{tabularx}

\usepackage{algorithm,algpseudocode}
 
\usepackage{amsfonts}

\DeclareMathOperator*{\argmin}{arg\,min}

\setcopyright{none}

\acmDOI{00.000/000_0}
 
\acmISBN{000-0000-00-000/00/00}
 
\acmConference[0000]{000}{0000}{Beijing, China} 
\acmYear{0000}
\copyrightyear{0000}
\acmPrice{00.00}

\begin{document}
	\title{Min-Max-Jump distance and its applications}
	
	\author{Gangli Liu}
	\affiliation{%
		\institution{Tsinghua University}
	}
	\email{gl-liu13@mails.tsinghua.edu.cn}

\begin{abstract}
We explore three applications of Min-Max-Jump distance (MMJ distance). MMJ-based K-means revises K-means with MMJ distance. MMJ-based Silhouette coefficient revises Silhouette coefficient  with MMJ distance. We also tested the Clustering with Neural Network and Index (CNNI) model with MMJ-based Silhouette coefficient. In the last application, we tested using Min-Max-Jump distance for predicting labels of new points, after a clustering analysis of data. Result shows Min-Max-Jump distance achieves good performances in all the three proposed applications. In addition, we devise several algorithms for calculating or estimating the distance.
\end{abstract}
 
	\maketitle
 
\section{Introduction}
Distance is a numerical measurement of how far apart objects or points are. It is usually formalized in mathematics using the notion of a metric space. A metric space is a set together with a notion of distance between its elements, usually called points. The distance is measured by a function called a metric or distance function.  Metric spaces are the most general setting for studying many of the concepts of mathematical analysis and geometry.

In this paper, we introduce three algorithms for calculating or estimating Min-Max-Jump distance (MMJ distance) and explore several applications of it. 
 
\section{RELATED WORK} \label{Sec_related}

Many distance measures have been proposed in literature, such as Euclidean distance or cosine similarity. These distance measures  often be found in algorithms like k-NN, UMAP, HDBSCAN, etc. The most common metric is Euclidean distance. Cosine similarity is often used as a way to counteract Euclidean distance’s problem in high dimensionality. The cosine similarity is the cosine of the angle between two vectors.

 Hamming distance is the number of values that are different between two vectors. It is typically used to compare two binary strings of equal length \cite{Li2009}.

 Manhattan distance is a geometry whose usual distance function or metric of Euclidean geometry is replaced by a new metric in which the distance between two points is the sum of the absolute differences of their Cartesian coordinates \cite{sinwar2014study}.

Chebyshev distance is defined as the greatest of difference between two vectors along any coordinate dimension \cite{coghetto2016chebyshev}. 

Minkowski distance or Minkowski metric is a metric in a normed vector space which can be considered as a generalization of both the Euclidean distance and the Manhattan distance \cite{groenen2001fuzzy}.

Jaccard index, also known as the Jaccard similarity coefficient, is a statistic used for gauging the similarity and diversity of sample sets \cite{fletcher2018comparing}.

Haversine distance is the distance between two points on a sphere given their longitudes and latitudes. It is similar to Euclidean distance in that it calculates the shortest path between two points. The main difference is that there is no straight line, since the assumption is that the two points are on a sphere \cite{chopde2013landmark}.

\section{ Definition of Min-Max-Jump}

\begin{table*}
	\caption{Table of notations}
	\begin{tabularx}{0.70\textwidth}{@{}XX@{}}
		\toprule
		$\Omega$ & A set of N points, with each point indexed from 1 to N;\\ \midrule
		$\Omega_{[1,n]}$ &The first $ n $ points of  $\Omega$,  indexed from 1 to n;\\ \midrule
		$\Omega_{n+1}$ &The $( n+1) $th point of  $\Omega$;\\\midrule
		$C_{i}$ &A cluster of points that is a subset of $\Omega$;\\\midrule
		$\xi_{i}$ &One-SCOM of $ C_i $;\\\midrule

		$\Omega$ + p &  Set $\Omega$ plus one new point $ p $. Since $ p \notin \Omega $, if $\Omega$ has N points, this new set now has $ N + 1 $ points;\\\midrule
		
		$\Psi_{(i,j,n,\Omega)}$  & $\Psi_{(i,j,n,\Omega)}$ is a sequence from point i to point j, which has length of n points. All the points in the sequence must belong to set $ \Omega $. That is to say, it is a path starts from i, and ends with j. For convenience, the path is not allowed to have loops, unless the start and the end is the same point;  \\\midrule

		$d(i,j)$  & $d(i,j)$ is a distance metric between pair of  points i and j, such as Euclidean distance;  \\\midrule
				
		$max\_jump(~\Psi_{(i,j,n,\Omega)}~)$	  & $max\_jump(~\Psi_{(i,j,n,\Omega)}~)$	 is the maximum jump in path $\Psi_{(i,j,n,\Omega)}$. A jump is the distance from two consecutive points p and q in the path; \\\midrule
		
		$\Theta_{(i,j,\Omega)}$	  & $\Theta_{(i,j,\Omega)}$	 is the set of all paths from point i to point j. A path in $\Theta_{(i,j,\Omega)}$ can have arbitrary number of points (at least two). All the points in a path must belong to set $ \Omega $; \\   \midrule
		
		$MMJ(i,j~|~\Omega)$ & 	 $MMJ(i,j~|~\Omega)$ is the MMJ distance between point i and j, where $ \Omega $ is the   $ \boldsymbol{Context} $ of the MMJ distance; \\ \midrule
	
		$ \mathbb{M}_{k,\Omega_{[1,k]}} $ & $ \mathbb{M}_{k,\Omega_{[1,k]}} $ is the pairwise MMJ distance matrix of $ \Omega_{[1,k]} $, which has shape $ k \times k $. The MMJ distances are under the $ \boldsymbol{Context} $ of $ \Omega_{[1,k]} $; \\ \midrule
		
		$ \mathbb{M}_{\Omega} $ & The pairwise MMJ distance matrix of $ \Omega $, $ \mathbb{M}_{\Omega}  = \mathbb{M}_{N,\Omega_{[1,N]}} $; \\ 
 
		\bottomrule
	\end{tabularx}
\label{tab:notations}
\end{table*}

\begin{definition}
Min-Max-Jump distance (MMJ distance)

$ \Omega $ is a set of points (at least one).  For any pair of points $p, q\in\Omega $, the distance between p and q is defined by a distance function d(p,q) (such as Euclidean distance). $i , j\in\Omega $, $\Psi_{(i,j,n,\Omega)}$ is a path from point i to point j, which has length of n points (see Table \ref{tab:notations}). $\Theta_{(i,j,\Omega)}$	 is the set of all paths from point i to point j. Therefore, $\Psi_{(i,j,n,\Omega)} \in \Theta_{(i,j,\Omega)}$. $max\_jump(~\Psi_{(i,j,n,\Omega)}~)$  is the maximum jump in path $\Psi_{(i,j,n,\Omega)}$. 

The Min-Max-Jump distance between a pair of points $ i,j $, which belong to $ \Omega $, is defined as:
	
\begin{equation}
\Pi =  \{max\_jump(\epsilon) ~ | ~ \epsilon \in \Theta_{(i,j,\Omega)}\}
\label{equ:mmj_defi1}
\end{equation}
	 
	\begin{equation}
 MMJ(i,j~|~\Omega) = min(\Pi)  
	\label{equ:mmj_defi2}
	\end{equation}
Where $ \epsilon $ is a path from point i to point j, $max\_jump(\epsilon)$  is the maximum jump in path $\epsilon$. $ \Pi $ is the set of all maximum jumps. ~ $min(\Pi) $ is the minimum of Set $ \Pi $.

Set $ \Omega $	is called the $ \boldsymbol{Context} $ of the Min-Max-Jump distance. 
It is easy to check $ MMJ(i,i~|~\Omega)  = 0$.
\label{def_MMJ}
 \end{definition}
In summary, Min-Max-Jump distance is the minimum of maximum jumps of all path between a pair of points, under the $ \boldsymbol{Context} $ of a set of points.

After publication of the first version of this paper, we received an email from Dr. Murphy, which refers a paper they published earlier \cite{Little2020}. After a preliminary examination, we conclude Definition \ref{def_MMJ} is actually the same thing as Definition 2.1 of \cite{Little2020}, which they called longest-leg path distance (LLPD). However, as stated by Dr. Murphy: ``this distance has actually been studied in many places in the literature, including in a paper released by myself and collaborators". It is a quite old distance that is referred to by many names in the literature. Including the maximum capacity path problem, the widest path problem, minimax path problem, and the bottleneck edge query problem \cite{pollack1960maximum,hu1961maximum,camerini1978min}.

There is a minor difference between Min-Max-Jump distance and other similar distances: Min-Max-Jump distance stresses the \emph{context} of the distance. The \emph{context} is like the condition in conditional probability. The difference becomes non-trivial when we need to calculate the pairwise MMJ distance matrix of a set $ S $, under the context of its superset $ X $, such as in Section 6.3 of \cite{liu2022clustering}. A set $ \Omega $ is a superset of another set $ B $ if all elements of the set $ B $ are elements of the set $ \Omega $. Another example is in Section \ref{sec:pred}, in Equation \ref{equ:g_i} and \ref{equ:g_i_new}, where we need to calculate the Min-Max-Jump distance between the same pair of points under different \emph{context}.

\subsection{An example} 
	\begin{figure} 
		\includegraphics[width=0.98\linewidth]{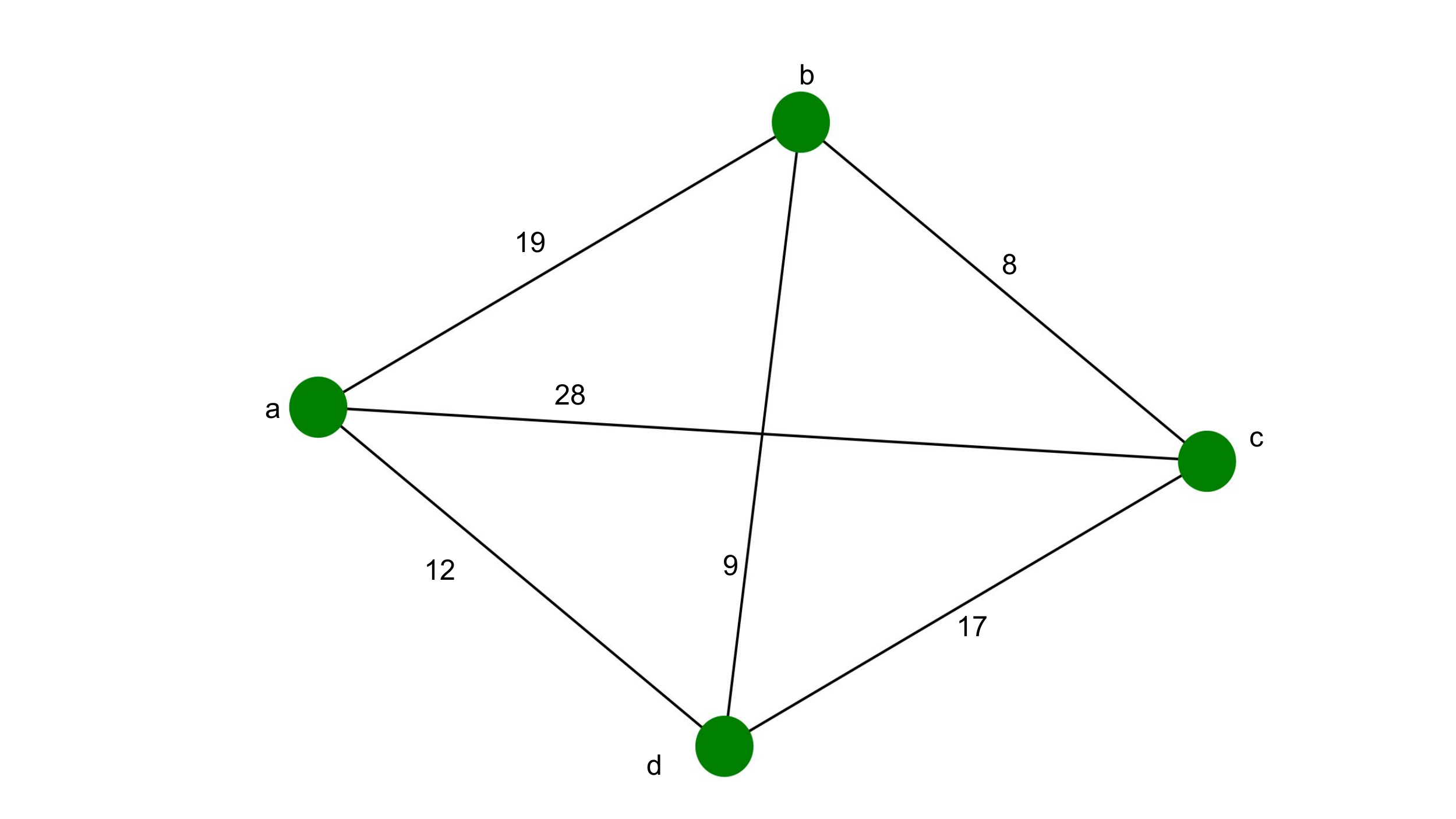}
	\caption{An example} \label{fig:an-example}
\end{figure}
Suppose Set $\Omega$ is composed of the four points in Figure \ref{fig:an-example}. There are five (non-looped) paths from point $ a $ to point $ c $ in Figure \ref{fig:an-example}:

\begin{enumerate}
	\item$  a \rightarrow c $, the maximum jump is 28;
	\item  $ a \rightarrow b \rightarrow c $, the maximum jump is 19;
	\item  $a \rightarrow  d \rightarrow c $, the maximum jump is 17;
	\item  $a \rightarrow b \rightarrow d \rightarrow  c $, the maximum jump is 19;	
	\item $ a \rightarrow d \rightarrow b \rightarrow c $, the maximum jump is 12.	
\end{enumerate}

According to Definition \ref{def_MMJ}, $ MMJ(a,c~|~\Omega)  = 12$.

To understand Min-Max-Jump distance, imagine someone is traveling by jumping in $ \Omega $. Suppose $ MMJ(i,j~|~\Omega)  = \delta $. If the person wants to reach $ j $ from $ i $, she must have the ability of jumping at least $ \delta $. Otherwise, $ j $ is unreachable from $ i $ for her. Whether the distance to a point is ``far" or ``near" is measured by how high (or how far) it requires a person to jump. If the requirement is high, then the point is ``far", otherwise, it is ``near".

\subsection{As a metric space} 
It can be easily checked that MMJ distance satisfies the four requirements as a metric. Suppose $ x, y, z \in \Omega$:

\begin{enumerate}
	\item  The distance from a point to itself is zero:	
		\begin{equation}
	MMJ(x,x~|~\Omega)  = 0
	\end{equation}
	
\item  	(Positivity) The distance between two distinct points is always positive:
		\begin{equation}
If ~ x \neq y, then ~   MMJ(x,y~|~\Omega)  >  0
\end{equation}

\item  	(Symmetry) The distance from x to y is always the same as the distance from y to x:
\begin{equation}
 MMJ(x,y~|~\Omega)  =  MMJ(y,x~|~\Omega)
\end{equation}

\item  The triangle inequality holds:
\begin{equation}
 MMJ(x,z~|~\Omega)  \leq  MMJ(x,y~|~\Omega) +  MMJ(y,z~|~\Omega)
\end{equation}

\end{enumerate}

\subsection{Other properties of MMJ distance} 

\begin{theorem} 
	\label{theorem1}
Suppose $ i,  j, p, q \in \Omega$
, 
\begin{equation}
MMJ(i,j~|~\Omega)  = \delta
\end{equation}
\begin{equation}
d(i,p) < \delta
\end{equation}
\begin{equation}
d(j,q) < \delta
\end{equation}
 then,
	\begin{equation}
MMJ(p,q~|~\Omega)  = \delta 
\end{equation}

\end{theorem}
where d(x,y) is a distance function (Table \ref{tab:notations}).

\begin{theorem} 
	\label{theorem2}
	Suppose $ r \in \{1,2, \dots, n \}$,
	
		\begin{equation}
	f(t) = max(  d(\Omega_{n+1},\Omega_t), ~ MMJ(\Omega_t,\Omega_r~|~\Omega_{[1,n]})  )  
	\end{equation}		
	\begin{equation}
	\mathbb{X} =  \{        f(t)   ~ | ~  t \in \{1,2, \dots, n \}   \}
	\end{equation}	
 then, 	
		\begin{equation}
MMJ(\Omega_{n+1},\Omega_r~|~\Omega_{[1,n+1]})  = min(\mathbb{X} )
	\end{equation}
 
\end{theorem}

For the meaning of $ \Omega_t,\Omega_r,\Omega_{[1,n]},   and ~  \Omega_{[1,n+1]}$, see Table \ref{tab:notations}.

\begin{corollary}
		\label{corollary1}
	Suppose $ r \in \{1,2, \dots, N \}, p \notin \Omega$,	
	\begin{equation}
	f(t) = max(  d(p,\Omega_t), ~ MMJ(\Omega_t,\Omega_r~|~\Omega)  )  
	\end{equation}		
	\begin{equation}
	\mathbb{X} =  \{        f(t)   ~ | ~  t \in \{1,2, \dots, N \}   \}
	\end{equation}		
	then,	
	\begin{equation}
	MMJ(p,\Omega_r~|~\Omega + p)  = min(\mathbb{X} )
	\end{equation}
\end{corollary}
For the meaning of $ \Omega + p $, see Table \ref{tab:notations}.

\begin{theorem} 
	\label{theorem3}
	Suppose $ i,j \in \{1,2, \dots, n \}$,	
	\begin{equation}
	x_1 = MMJ(\Omega_{i},\Omega_j~|~\Omega_{[1,n]})
	\end{equation}		
	\begin{equation}
	t_1 = MMJ(\Omega_{n+1},\Omega_i~|~\Omega_{[1,n+1]})
	\end{equation}		
	\begin{equation}
	t_2 = MMJ(\Omega_{n+1},\Omega_j~|~\Omega_{[1,n+1]})
\end{equation}	
	\begin{equation}
x_2 =max(t_1 , ~  t_2)
\end{equation}		
	then,	
	\begin{equation}
	MMJ(\Omega_{i},\Omega_j~|~\Omega_{[1,n+1]})  = min(x_1 , ~  x_2)
	\end{equation}
	
\end{theorem}

Since it is easy to prove the theorems and corollary, here we omit the proofs.

\section{ Calculation of Min-Max-Jump} 
We propose three methods to calculate or estimate Min-Max-Jump distance. For other methods see Section 2.3.3. of \cite{Little2020}.
 
\subsection{ MMJ distance by recursion} 
This section presents an algorithm for calculating the $ \mathbb{M}_{\Omega} $. $ \mathbb{M}_{\Omega} $ is the pairwise MMJ distance matrix of $ \Omega $ (Table \ref{tab:notations}). $ \mathbb{M}_{k,\Omega_{[1,k]}} $ is the MMJ distance matrix of the first k points of $\Omega$ (Table \ref{tab:notations}). Note $ \mathbb{M}_{2,\Omega_{[1,2]}} $ is simple to calculate. $ \mathbb{M}_{\Omega}  = \mathbb{M}_{N,\Omega_{[1,N]}} $. $\mathbb{M}_{\Omega} $ is a  $ N \times N $ symmetric matrix. Rows and columns of $ \mathbb{M}_{\Omega} $ are indexed from 1 to N.

\begin{algorithm}[H] 
	\caption{MMJ distance by recursion}
	\begin{algorithmic}[1]
		\Require{$\Omega$} 
		\Ensure{$ \mathbb{M}_{\Omega} $}
		\Statex
		\Function{MMJ\_by\_recursion}{$\Omega$}
		\State {$N$ $\gets$ {$length(\Omega)$}}
		\State {Initialize $ \mathbb{M}_{\Omega} $ with zeros}
		\State {Calculate $ \mathbb{M}_{2,\Omega_{[1,2]}} $, fill in $ \mathbb{M}_{\Omega}[1,2] $ and $ \mathbb{M}_{\Omega}[2,1] $ }

		\For{$n \gets 3 $ to $N$}        
			\For{$r \gets 1 $ to $n -1$}  
			\State {Calculate $MMJ(\Omega_{n},\Omega_r~|~\Omega_{[1,n]})$, fill in $ \mathbb{M}_{\Omega}[n,r] $ and $ \mathbb{M}_{\Omega}[r,n] $}
			\EndFor
		
		\For{$i \gets 1 $ to $n -1$}  
		\For{$j \gets 1 $ to $n -1$}  
		\If{$ i < j $}
		\State {Calculate $MMJ(\Omega_{i},\Omega_j~|~\Omega_{[1,n]})$, update $ \mathbb{M}_{\Omega}[i,j] $ and $ \mathbb{M}_{\Omega}[j,i] $}
		\EndIf		
		\EndFor
		\EndFor

		\EndFor
		\State \Return {$\mathbb{M}_{\Omega} $}
		\EndFunction
	\end{algorithmic}
	\label{alg:recursion}
\end{algorithm}
Step 7 of Algorithm \ref{alg:recursion} can be calculated with the conclusion of Theorem \ref{theorem2}; Step 12 of Algorithm \ref{alg:recursion} can be calculated with the conclusion of Theorem \ref{theorem3}.

Algorithm \ref{alg:recursion} has complexity of $\mathcal{O}(n^3)$, where n is the cardinality of Set $\Omega$.

\subsection{ MMJ distance by estimation and copy} 
In last section, we calculate $ \mathbb{M}_{\Omega} $ precisely by recursion. In this section, we provide a way for calculating $ \mathbb{M}_{\Omega} $ approximately.

Since Min-Max-Jump distance is the minimum of maximum jumps of all path from point i to j, we can sample some paths from point i to j, calculate the maximum jump of each path, then calculate the minimum of the maximum jumps, and use the minimum as an approximation of $ MMJ(i,j~|~\Omega) $.

Once we obtained $ MMJ(i,j~|~\Omega) $, according to Theorem \ref{theorem1}, $ MMJ(p,q~|~\Omega)   =   MMJ(i,j~|~\Omega) $, if $ p,q $ are near $ i,j $ respectively. Therefore, we can copy the result of $  MMJ(i,j~|~\Omega)  $ to $ MMJ(p,q~|~\Omega) $. Further, we can copy the result to another pair of points $ m,n $, which are near $ p,q $, respectively. This copying process can be conducted recursively until all the pairs of near points are processed.

\begin{algorithm}[H] 
	\caption{MMJ distance by Estimation and Copy}
	\label{alg:estima}
	\begin{algorithmic}[1]
		\Require{$\Omega$} 
		\Ensure{$ \mathbb{M}_{\Omega} $}
		\Statex
		\Function{MMJ\_by\_estimation}{$\Omega$}
		\State {$N$ $\gets$ {$length(\Omega)$}}
		\State {Initialize $ \mathbb{M}_{\Omega} $ with zeros}

		\For{$i \gets 1 $ to $N$}  
		\For{$j \gets 1 $ to $N$}  
		\If{$ i < j $}
	\State {Sample some paths from point $ \Omega_{i} $ to $ \Omega_j $, estimate $MMJ(\Omega_{i},\Omega_j~|~\Omega)$ , update $ \mathbb{M}_{\Omega}[i,j] $ and $ \mathbb{M}_{\Omega}[j,i] $}
				
		\State {Recursively copy $MMJ(\Omega_{i},\Omega_j~|~\Omega)$ to $ \mathbb{M}_{\Omega}[p,q] $ and $ \mathbb{M}_{\Omega}[q,p] $, where $ \Omega_p, \Omega_q $ are pair of near points to $ \Omega_i, \Omega_j $, respectively}
		\EndIf		
		\EndFor
		\EndFor
 
		\State \Return {$\mathbb{M}_{\Omega} $}
		\EndFunction
	\end{algorithmic}
\end{algorithm}

\begin{figure*} 
	\begin{subfigure}{0.24\textwidth}
		\includegraphics[width=\linewidth]{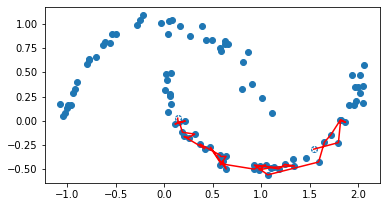}

	\end{subfigure}    
	\begin{subfigure}{0.24\textwidth}
		\includegraphics[width=\linewidth]{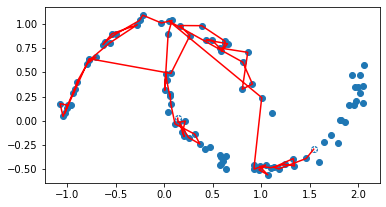}
  
	\end{subfigure}    
	\begin{subfigure}{0.24\textwidth}
	\includegraphics[width=\linewidth]{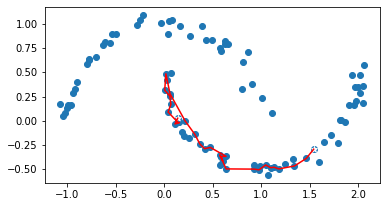}
 
\end{subfigure}    
	\begin{subfigure}{0.24\textwidth}
	\includegraphics[width=\linewidth]{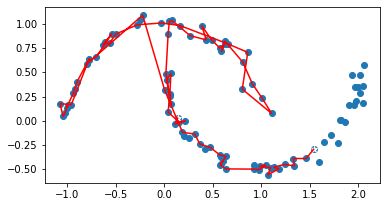}
 
\end{subfigure}  

	\caption{Four sampled paths between two points} \label{fig:4-paths}
\end{figure*}

To obtain an estimation of $MMJ(\Omega_{i},\Omega_j~|~\Omega)$ in Step 7 of Algorithm \ref{alg:estima}, we can run Algorithm \ref{alg:p_to_p} for k times, then calculate the minimum of the $ max\_jumps $. E.g., Figure \ref{fig:4-paths} are four sampled paths between two points by Algorithm \ref{alg:p_to_p}.

\begin{algorithm}[H] 
	\caption{Sample one path by stochastic greedy}
	\label{alg:p_to_p}
	\begin{algorithmic}[1]
		\Require{$\Omega$, start,end} 
		\Ensure{maximum jump of one sampled path}
		\Statex
		\Function{maximum\_jump\_one\_path}{$\Omega$, start, end}		
				\If{$ start = end $}
		\State \Return {$0 $} 
		\EndIf			
			\State {$N$ $\gets$ {$length(\Omega)$}}
		\State {$ path\_list \gets empty\_list$}		
		\State {$ remaining\_list \gets range(1~ to ~N)$}
			\State {$ next \gets start $}			
				\State {$ remaining\_list.remove(next) $}				
					\State {$ path\_list.append(next)$}					
					\While{$next ~~ \neq ~end$} 					
						\State {Find k near points to $ next $ (e.g., k = 3) from the $ remaining\_list $, noted $ near\_points$. If the $ remaining\_list $ has less than k points, then put them all to $ near\_points$}, {$ nearest\_point = near\_points[0]$}											
								\If{$end ~~ = nearest\_point$}
						\State {$ next \gets end $}
						\Else
						\State {$ next \gets random.choice(near\_points) $}
						\EndIf
						\State {$ remaining\_list.remove(next) $}					
					\State {$ path\_list.append(next)$}								
					\EndWhile					
					\State {Find out the maximum jump in $  path\_list $, noted $ max\_jump $}			
		\State \Return {$ max\_jump $}
		\EndFunction
	\end{algorithmic}
\end{algorithm}

The merit of  the ``estimation and copy" method is that the computation can be accelerated with parallel computing. The demerit is that we only get an approximation of $\mathbb{M}_{\Omega} $.

\subsection{ MMJ distance by calculation and copy} 
In last section, we estimate a  MMJ distance and copy it to other positions in $ \mathbb{M}_{\Omega} $. In this section, we precisely calculate a  MMJ distance and copy it to other positions in $ \mathbb{M}_{\Omega} $.

A well-known fact about MMJ distance is: ``the path between any two nodes in a minimum spanning tree (MST) is a minimax path". A minimax path in an undirected graph is a path between two vertices v, w that minimizes the maximum weight of the edges on the path. That is to say, it is a MMJ path. By utilizing this fact, we propose Algorithm \ref{alg:calcopy}.
 
\begin{algorithm}[H] 
	\caption{MMJ distance by Calculation and Copy}
	\label{alg:calcopy}
	\begin{algorithmic}[1]
		\Require{$\Omega$} 
		\Ensure{$ \mathbb{M}_{\Omega} $}
		\Statex
		\Function{MMJ\_Calculation\_and\_Copy}{$\Omega$}
		\State {Initialize $ \mathbb{M}_{\Omega} $ with zeros}
		\State {Construct a MST of  $ \Omega $,  noted $ T $}
		\State {Sort edges of $ T $ from large to small, generate a list, noted $ L $}
		\For{e  in $ L $}
		\State {Remove $ e $ from $ T $. It will result in two connected sub-trees, $ T_1 $ and $ T_2 $;}
		\State {For all pair of nodes $ (p,q) $, where $p \in  T_1$, $q \in  T_2$. Fill in $ \mathbb{M}_{\Omega}[p, q] $ and $ \mathbb{M}_{\Omega}[q,p] $ with $ e $.}
		\EndFor
		
		\State \Return {$\mathbb{M}_{\Omega} $}
		\EndFunction
	\end{algorithmic}
\end{algorithm}

The complexity  of Algorithm \ref{alg:calcopy} is $\mathcal{O}(n^2)$. Because the construction of a MST of a complete graph is $\mathcal{O}(n^2)$. During the ``for" part (Step 5 to 8) of the algorithm, it accesses each cell of $ \mathbb{M}_{\Omega} $ only once. Unlike Algorithm \ref{alg:recursion}, which accesses each cell of $ \mathbb{M}_{\Omega} $ for $\mathcal{O}(n)$ times .

\section{Applications of Min-Max-Jump} 
This section we explore several applications of Min-Max-Jump distance, and test the applications with experiments. All the Min-Max-Jump distances in the experiments are calculated with Algorithm \ref{alg:recursion}.
 
\subsection{MMJ-based K-means}  
K-means clustering aims to partition n observations into k clusters in which each observation belongs to the cluster with the nearest mean (cluster center or cluster centroid), serving as a prototype of the cluster \cite{bock2007clustering}.   Standard K-means uses Euclidean distance. We can revise K-means to use Min-Max-Jump distance, with the  cluster centroid replaced by the Semantic Center of Mass (SCOM) (particularly, One-SCOM) of each cluster. For the definition of SCOM, see a previous paper \cite{liu2021topic}. One-SCOM is like medoid, but has some difference from medoid. Section 6.3 of \cite{liu2021topic} compares One-SCOM and medoid. In simple terms, One-SCOM of a set of points, is the point which has the  smallest sum of squared distances to all points in the set.

\begin{figure*} 
	\begin{subfigure}{0.2\textwidth}
		\includegraphics[width=\linewidth]{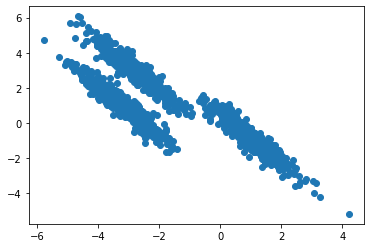}
		\caption{data A}  
	\end{subfigure}    
	\begin{subfigure}{0.2\textwidth}
		\includegraphics[width=\linewidth]{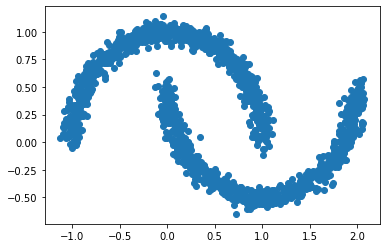}
		\caption{data B}  
	\end{subfigure}    
	\begin{subfigure}{0.2\textwidth}
		\includegraphics[width=\linewidth]{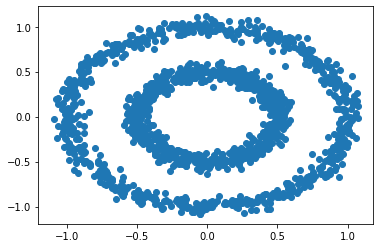}
		\caption{data C}  
	\end{subfigure}    

	\begin{subfigure}{0.2\textwidth}
	\includegraphics[width=\linewidth]{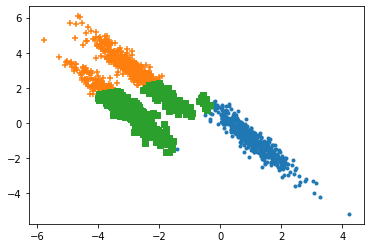}
	\caption{data A, Standard K-means}  
\end{subfigure}    
\begin{subfigure}{0.2\textwidth}
	\includegraphics[width=\linewidth]{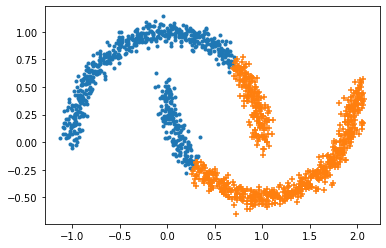}
	\caption{data B, Standard K-means}  
\end{subfigure}    
\begin{subfigure}{0.2\textwidth}
	\includegraphics[width=\linewidth]{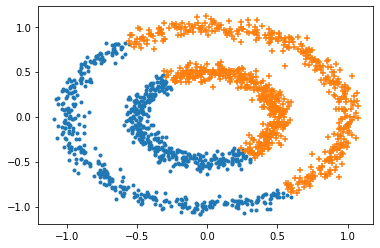}
	\caption{data C, Standard K-means}  
\end{subfigure} 

	\begin{subfigure}{0.2\textwidth}
	\includegraphics[width=\linewidth]{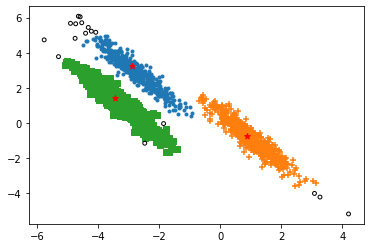}
	\caption{data A, MMJ-K-means}  
\end{subfigure}    
\begin{subfigure}{0.2\textwidth}
	\includegraphics[width=\linewidth]{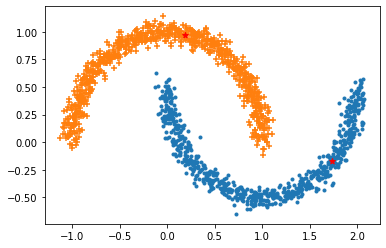}
	\caption{data B, MMJ-K-means}  
\end{subfigure}    
\begin{subfigure}{0.2\textwidth}
	\includegraphics[width=\linewidth]{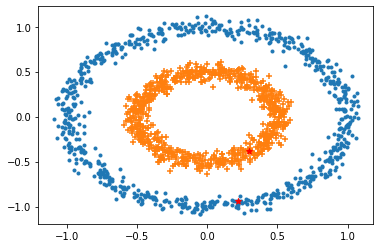}
	\caption{data C, MMJ-K-means}  
\end{subfigure} 
 
	\caption{Standard K-means vs. MMJ-K-means} \label{fig:MMJ-K-means}
\end{figure*}

\begin{figure*} 
	\begin{subfigure}{0.24\textwidth}
	\includegraphics[width=\linewidth]{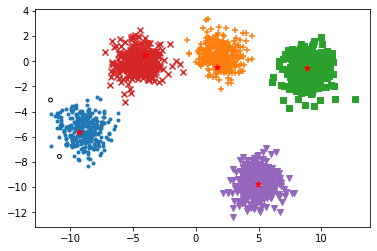}
	\caption{data 1}  
\end{subfigure}    
	\begin{subfigure}{0.24\textwidth}
		\includegraphics[width=\linewidth]{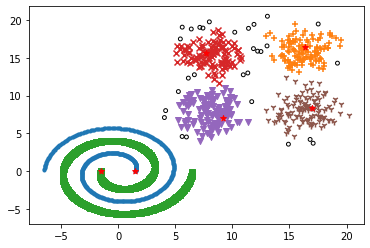}
		\caption{data 54}  
	\end{subfigure}    
	\begin{subfigure}{0.24\textwidth}
		\includegraphics[width=\linewidth]{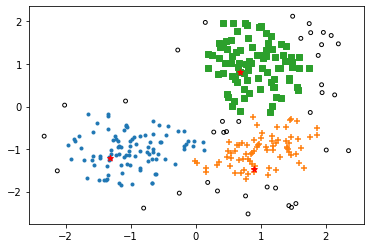}
		\caption{data 62}  
	\end{subfigure}    
	\begin{subfigure}{0.24\textwidth}
	\includegraphics[width=\linewidth]{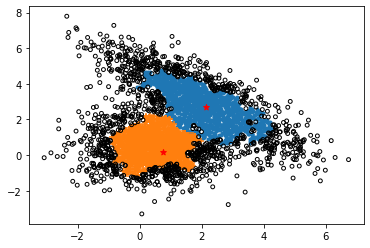}
	\caption{data 76}  
\end{subfigure}

	\begin{subfigure}{0.24\textwidth}
	\includegraphics[width=\linewidth]{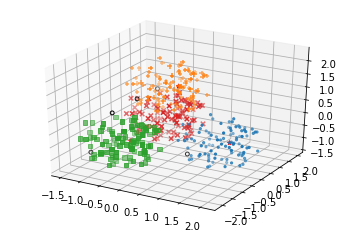}
	\caption{data 78}  
\end{subfigure}    
\begin{subfigure}{0.24\textwidth}
	\includegraphics[width=\linewidth]{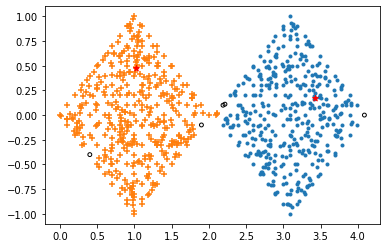}
	\caption{data 83}  
\end{subfigure}    
\begin{subfigure}{0.24\textwidth}
	\includegraphics[width=\linewidth]{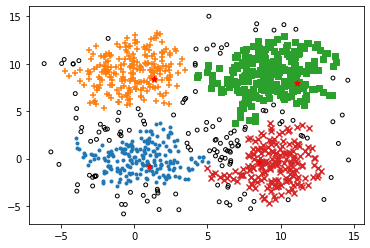}
	\caption{data 89}  
\end{subfigure}    
\begin{subfigure}{0.24\textwidth}
	\includegraphics[width=\linewidth]{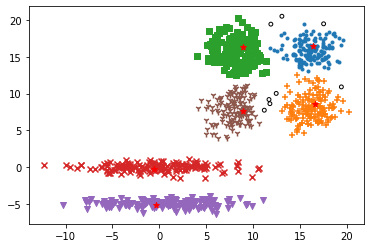}
	\caption{data 102}  
\end{subfigure} 
 
	\caption{Eight more samples of MMJ-K-means} \label{fig:big_8}
\end{figure*}

Standard K-means usually cannot deal with non-sphere shaped data. MMJ-based K-means (MMJ-K-means) can cluster non-sphere shaped data.  Figure \ref{fig:MMJ-K-means} compares Standard K-means and MMJ-K-means, on clustering three data which come from the scikit-learn project \cite{scikit-learn}. Figure \ref{fig:big_8} are eight more samples of MMJ-K-means. The data sources corresponding to the data IDs can be found at this URL. \footnote{\url{https://github.com/mike-liuliu/Min-Max-Jump-distance}}
 
It can be seen MMJ-K-means can (almost) work properly for clustering the 11 data, which have different kinds of shapes. The black circles are Border points (Definition \ref{def_border_p}), the red stars are the center (One-SCOM) of each cluster. During training of MMJ-K-means, the Border points are randomly allocated to one of its nearest centers.
\begin{definition}
	Border point
	
	A point is defined to be a Border point if its nearest mean (center or One-SCOM) is not unique.	
	\label{def_border_p}
\end{definition}

After analysis of MMJ-K-means, if we want to re-allocate a label to a  Border point, we can use the method discussed in Section \ref{sec:pred}, deeming them as new points and process them with Equation \ref{equ:g_i}, \ref{equ:Y2}, and \ref{equ:arg_min_y}. Or just considering them as outliers and do nothing.

\subsection{MMJ-based Silhouette coefficient}  
Silhouette coefficient is a metric used to calculate the goodness of a clustering technique. 

The Silhouette coefficient for a single sample is given as:
\begin{equation*}
s = \frac{b - a}{max(a, b)}
\label{eq:S_index}
\end{equation*}
where $ a $ is the mean distance between a sample and all other points in the same class. $ b $ is the mean distance between a sample and all other points in the next nearest cluster. The Silhouette coefficient for a set of samples is given as the mean of Silhouette coefficient for each sample. 

We can also revise Silhouette coefficient to use Min-Max-Jump distance, forming a new internal clustering evaluation index called MMJ-based Silhouette coefficient (MMJ-SC). We tested the performance of MMJ-SC with the 145 data sets mentioned in \cite{liu2022new}.
MMJ-SC obtained a good performance score compared with the other seven internal clustering evaluation indices  mentioned in \cite{liu2022new}. Readers can check Table \ref{tab:Accu} and compare with Table 5 of \cite{liu2022new}. 

MMJ-based Calinski-Harabasz index (MMJ-CH) and MMJ-based Davies-Bouldin index (MMJ-DB) are also tested. In calculation of these two indices, the center/centroid of a cluster is replaced by the One-SCOM of the cluster again, as in MMJ-K-means. It can be seen that MMJ distance systematically improves the three internal clustering evaluation indices, and outperforms other indices like DBCV \cite{moulavi2014density}, CDbw \cite{halkidi2008density}, and  VIASCKDE \cite{csenol2022viasckde} (Table \ref{tab:Accu}).

\begin{table*}
	\begin{center}
		\scalebox{0.9}{
			
			\begin{tabular}{lllllllllll}
				\toprule
				{}    &  CH   & SC    &  DB   &  CDbw   &   DBCV   &   VIASCKDE   & New & MMJ-SC & MMJ-CH & MMJ-DB\\
				\midrule
				Accuracy & 27/145 & 38/145 & 42/145  & 8/145  & 56/145  & 11/145 & 74/145 &   83/145 &   90/145 &   69/145 \\
				\bottomrule
		\end{tabular} 		}		
		
		\caption{Accuracy of the ten indices}
		\label{tab:Accu}
	\end{center}
\end{table*}

\subsubsection{Using MMJ-SC in CNNI}  
\begin{figure} 
	\begin{subfigure}{0.35\textwidth}
		\includegraphics[width=\linewidth]{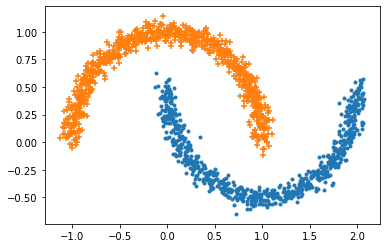}
	\end{subfigure}    

	\begin{subfigure}{0.35\textwidth}
		\includegraphics[width=\linewidth]{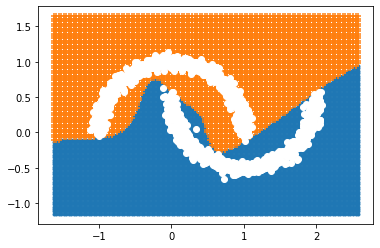}  
	\end{subfigure}    

	\caption{Clustering result and decision boundary of data B by CNNI using MMJ-SC} \label{fig:data_b_clu}
\end{figure}

The Clustering with Neural Network and Index (CNNI) model uses a Neural Network to cluster data points. Training of the Neural Network mimics supervised learning, with an internal clustering evaluation index acting as the loss function \cite{liu2022clustering}. CNNI with 
standard Silhouette coefficient as the internal clustering evaluation index, seems cannot deal with non-convex shaped data, such as data B and data C in Figure \ref{fig:MMJ-K-means}. MMJ-SC provides a new internal clustering evaluation index for CNNI, which can deal with non-convex shaped data. E.g., Figure \ref{fig:data_b_clu} is the clustering result and decision boundary of data B by CNNI using MMJ-SC. It uses Neural Network C of the CNNI paper \cite{liu2022clustering}.

\subsection{MMJ classifier} \label{sec:pred}
	\begin{figure} 
	\begin{subfigure}{0.35\textwidth}
		\includegraphics[width=\linewidth]{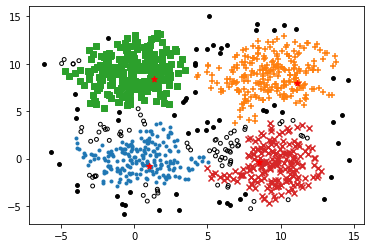}
		\caption{Weak and strong border points}
	\end{subfigure}    

	\begin{subfigure}{0.35\textwidth}
		\includegraphics[width=\linewidth]{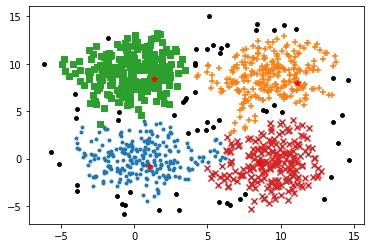}  
		\caption{Strong border points only}
	\end{subfigure}    

	\caption{Differentiating border points} \label{fig:strong_weak}
\end{figure}
	\begin{figure*} 
	\begin{subfigure}{0.28\textwidth}
	\includegraphics[width=\linewidth]{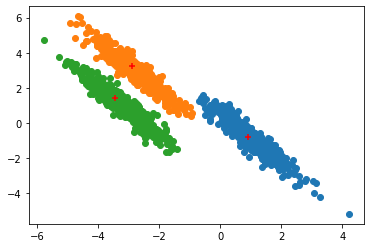}
	\caption{ Clusters and One-SCOMs} 
\end{subfigure}   
\begin{subfigure}{0.28\textwidth}
	\includegraphics[width=\linewidth]{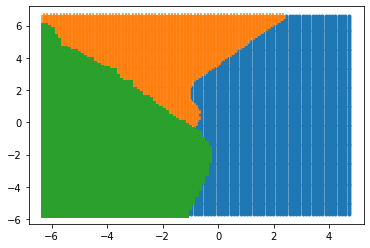}  
	\caption{ Decision boundary} 
\end{subfigure}    
\begin{subfigure}{0.28\textwidth}
	\includegraphics[width=\linewidth]{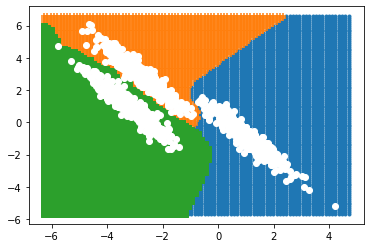} 
	\caption{Decision boundary and data $ A $} 
\end{subfigure} 

	\begin{subfigure}{0.28\textwidth}
		\includegraphics[width=\linewidth]{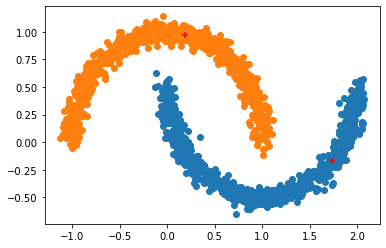}
			\caption{ Clusters and One-SCOMs} 
	\end{subfigure}   
	\begin{subfigure}{0.28\textwidth}
		\includegraphics[width=\linewidth]{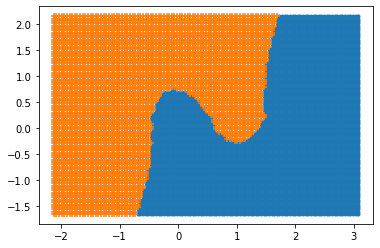}  
		\caption{ Decision boundary} 
	\end{subfigure}    
	\begin{subfigure}{0.28\textwidth}
	\includegraphics[width=\linewidth]{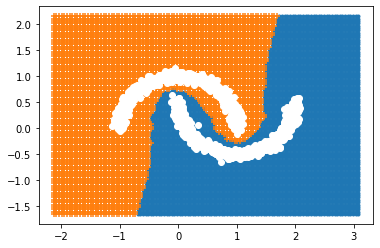} 
		\caption{Decision boundary and data $ B $} 
\end{subfigure} 

	\begin{subfigure}{0.28\textwidth}
	\includegraphics[width=\linewidth]{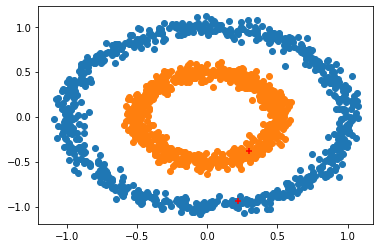}
	\caption{ Clusters and One-SCOMs} 
\end{subfigure}   
\begin{subfigure}{0.28\textwidth}
	\includegraphics[width=\linewidth]{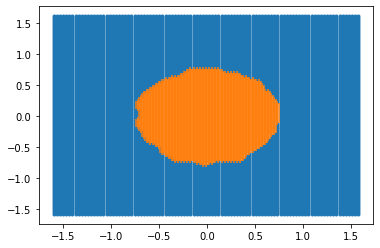}  
	\caption{ Decision boundary} 
\end{subfigure}    
\begin{subfigure}{0.28\textwidth}
	\includegraphics[width=\linewidth]{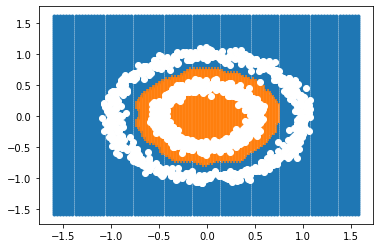} 
	\caption{Decision boundary and data $ C $} 
\end{subfigure}

	\caption{Predicting labels of new points} \label{fig:predict}
\end{figure*}
	\begin{figure*} 
	\begin{subfigure}{0.3\textwidth}
	\includegraphics[width=\linewidth]{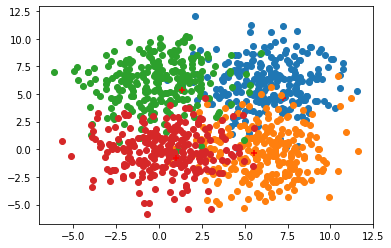}
	\caption{A not well separated clustering}  
\end{subfigure}    
	\begin{subfigure}{0.3\textwidth}
		\includegraphics[width=\linewidth]{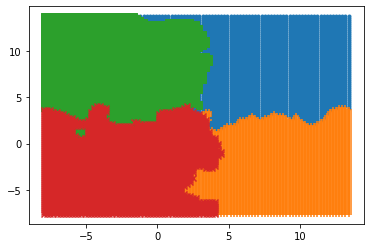}
		\caption{Decision boundary, MMJ}  
	\end{subfigure}    
	\begin{subfigure}{0.3\textwidth}
		\includegraphics[width=\linewidth]{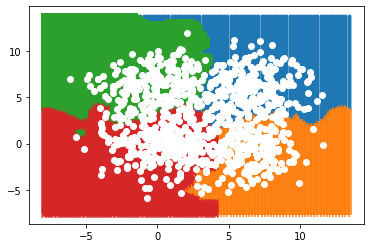}
		\caption{Decision boundary and data, MMJ}  
	\end{subfigure}

	\begin{subfigure}{0.3\textwidth}
	\includegraphics[width=\linewidth]{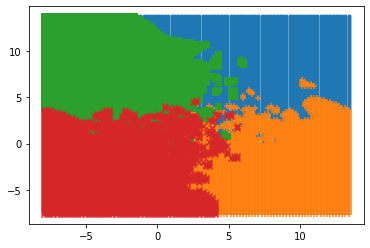}
	\caption{Decision boundary, k-NN}  
\end{subfigure}    
\begin{subfigure}{0.3\textwidth}
	\includegraphics[width=\linewidth]{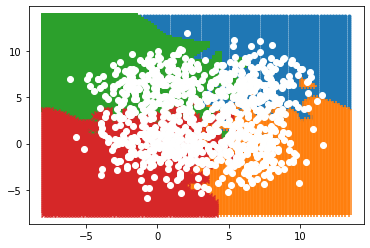}
	\caption{Decision boundary and data, k-NN}  
\end{subfigure}

	\caption{Comparing MMJ classifier and k-NN (k = 1)} \label{fig:MMJ-com-knn}
\end{figure*}

	\begin{figure*} 
	\begin{subfigure}{0.3\textwidth}
	\includegraphics[width=\linewidth]{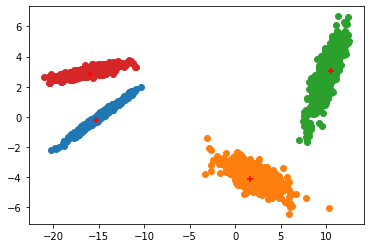}
	\caption{Clusters and One-SCOMs}  
\end{subfigure}    
	\begin{subfigure}{0.3\textwidth}
		\includegraphics[width=\linewidth]{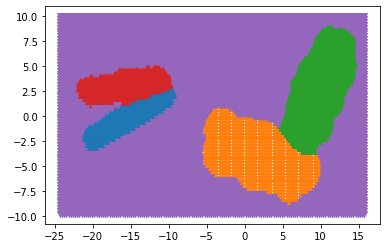}
		\caption{Decision boundary}  
	\end{subfigure}    
	\begin{subfigure}{0.3\textwidth}
		\includegraphics[width=\linewidth]{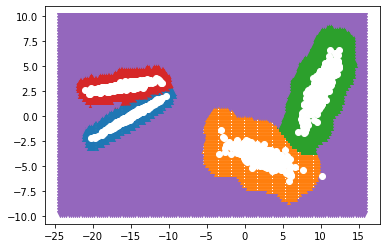}
		\caption{Decision boundary and data}  
	\end{subfigure}

	\caption{Border points are colored with the background color} \label{fig:MMJ-border}
\end{figure*}

Non-parametric clustering models like Spectral clustering and Hierarchical clustering can deal with non-convex shaped data. However, they are non-parametric. That means we cannot get a set of parameters, and predict the label of a new point based on the parameters. 

To predict label of a new point, we may need to train a supervised model with the clustering result, then predict the label with the supervised model. MMJ distance provides a new method for predicting labels of new points.

Suppose a data set $ \Omega $ is clustered into K clusters, with each cluster noted $ C_i $, $  i \in \{1,2, \dots, K\} $, the One-SCOM of $ C_i $ is $ \xi_i $. $ \xi_i $ is calculated based on $ \mathbb{M}_{C_i} $, the pairwise MMJ distance matrix of $C_i $.

Suppose we got a new point $ p, ~p \notin  \Omega $, we want to predict the label of the new point, based on the clustering result of $ \Omega $. Label of $ p $ is noted $ L(p) $.

	\begin{equation}
g(i) =    MMJ( p , \xi_i~|~C_i + p)  
\label{equ:g_i}
\end{equation}		
\begin{equation}
\mathbb{Y} =  \{      g(i)   ~ | ~  i \in \{1,2, \dots, K\}   \}
\label{equ:Y2}
\end{equation}		
then,	
\begin{equation}
L(p)  = \argmin(~ Y ~)
\label{equ:arg_min_y}
\end{equation}

$ g(i) $ can be calculated with the conclusion of Corollary \ref{corollary1}.

Figure \ref{fig:predict} illustrates decision boundaries of predicting labels of 10,000 new points, compared with the data. Figure \ref{fig:MMJ-com-knn} is an example illustrating difference between k-nearest neighbors (k-NN) and Min-Max-Jump distance to predict labels of new points, when the clusters are not well separated. It can be seen MMJ classifier has done a good job for predicting labels of new points.

Another choice is to replace Equation \ref{equ:g_i} with Equation \ref{equ:g_i_new}. Note this time $ \xi_i $ is calculated based on $ \mathbb{M}_{\Omega} $, not $ \mathbb{M}_{C_i} $. In this revision, a new point is likely to become the Border point (Definition \ref{def_border_p}) of multiple clusters, because its MMJ distance to the centers (One-SCOMs) of  these clusters are the same. Figure \ref{fig:MMJ-border} is result of using Equation \ref{equ:g_i_new}. The Border points are colored with the background color. The interesting part is, as shown by the figure, we can get an `envelope' for each cluster.
	\begin{equation}
g(i) =    MMJ( p , \xi_i~|~\Omega + p)  
\label{equ:g_i_new}
\end{equation}

The application discussed in this section and MMJ-based K-means facilitates a new clustering model, for dealing with arbitrarily-shaped large data. We first get a small sample of the data, cluster it with MMJ-based K-means, then predict labels of the remaining data points. 

Another choice is to use the CNNI model to analyze the small sample, then predict labels of the remaining data points. Since CNNI is parametric model, it is effortless for it to predict labels of remaining points.

\section{In a directed graph} 
Algorithm \ref{alg:recursion} can be revised to solve the minimax path problem or bottleneck shortest path problem in a directed graph.

To calculate the MMJ distance  in a directed graph, we revise Step 7 of Algorithm \ref{alg:recursion} to:

\emph{Calculate $MMJ(\Omega_{n},\Omega_r~|~\Omega_{[1,n]})$ with Theorem \ref{theorem2}, fill in $ \mathbb{M}_{\Omega}[n,r] $; Calculate $MMJ(\Omega_r,\Omega_n~|~\Omega_{[1,n]})$ with Theorem \ref{theo_2_back}, fill in  $ \mathbb{M}_{\Omega}[r,n] $.}

And revise Step 12 of Algorithm \ref{alg:recursion} to:

\emph{Calculate $MMJ(\Omega_i,\Omega_j~|~\Omega_{[1,n]})$ and $MMJ(\Omega_j,\Omega_i~|~\Omega_{[1,n]})$ with Theorem \ref{theo_3_back}, update $ \mathbb{M}_{\Omega}[i,j] $ and $ \mathbb{M}_{\Omega}[j,i] $.}

Note the orders of distances in the new theorems have been changed. Since in a directed graph, the pairwise distance matrix is not symmetric anymore; direction of distances matters now.

\begin{theorem} 
	\label{theo_2_back}
	Suppose $ r \in \{1,2, \dots, n \}$,
	
	\begin{equation}
	f(t) = max(MMJ(\Omega_r,\Omega_t~|~\Omega_{[1,n]}), ~  d(\Omega_t, \Omega_{n+1}))  
	\end{equation}		
	\begin{equation}
	\mathbb{X} =  \{        f(t)   ~ | ~  t \in \{1,2, \dots, n \}   \}
	\end{equation}	
	then, 	
	\begin{equation}
	MMJ(\Omega_r, \Omega_{n+1}~|~\Omega_{[1,n+1]})  = min(\mathbb{X} )
	\end{equation}
	
\end{theorem}

\begin{theorem} 
	\label{theo_3_back}
	Suppose $ i,j \in \{1,2, \dots, n \}$,	
	\begin{equation}
	x_1 = MMJ(\Omega_{i},\Omega_j~|~\Omega_{[1,n]})
	\end{equation}		
	\begin{equation}
	t_1 = MMJ(\Omega_i, \Omega_{n+1}~|~\Omega_{[1,n+1]})
	\end{equation}		
	\begin{equation}
	t_2 = MMJ(\Omega_{n+1},\Omega_j~|~\Omega_{[1,n+1]})
	\end{equation}	
	\begin{equation}
	x_2 =max(t_1 , ~  t_2)
	\end{equation}		
	then,	
	\begin{equation}
	MMJ(\Omega_{i},\Omega_j~|~\Omega_{[1,n+1]})  = min(x_1 , ~  x_2)
	\end{equation}
	
\end{theorem}

\section{Solving the widest path problem}
The widest path problem (maximum capacity path problem) is the problem of finding a path between two designated vertices in a weighted graph, maximizing the weight of the minimum-weight edge in the path. It is a ``mirror'' problem of the minimax path problem.

Algorithm \ref{alg:recursion} can be revised to solve the widest path problem, in both directed and undirected graphs. Three revisions are needed to solve the problem:

\begin{enumerate}
	\item Revise the base distance, $d(i, j)$. The capacity of a node to itself is infinity, that is: $d(i, i) = \infty$. When two nodes are not directly connected, their direct capacity is zero, that is: $d(i, j) = 0$, if $ i $ and $ j $ are not directly connected.
	\item Initialize with infinity. In step 3 of Algorithm \ref{alg:recursion}, we initialize $ \mathbb{M}_{\Omega} $ with infinity.
	\item Swap the \emph{min} and \emph{max} operators. We swap the \emph{min} and \emph{max} operators in Theorem \ref{theorem2},  \ref{theorem3}, \ref{theo_2_back}, and \ref{theo_3_back}.
\end{enumerate}

Algorithm \ref{alg:calcopy} can also be revised to solve the widest path problem in undirected graphs, by constructing a maximum spanning tree and sort the edges from small to large.

\section{Discussion} 
\subsection{Using PAM} 
Since One-SCOM is like medoid, in MMJ-K-means, we can also use the Partitioning Around Medoids (PAM)  algorithm or its variants to find the One-SCOMs \cite{schubert2021fast}.

\subsection{Multiple One-SCOMs in one cluster} 
There might be multiple One-SCOMs which have the same smallest sum of squared distances to all points in the cluster. Usually they are not far from each other. We can arbitrarily choose one or keep them all. If we keep them all, when calculating a point's MMJ distance to the One-SCOMs of a cluster, e.g., in Equation \ref{equ:g_i} and \ref{equ:g_i_new}, then we can select the minimum of the MMJ distances.

\subsection{Differentiating border points}
Border points defined in  Definition \ref{def_border_p} can further be  differentiated as weak and strong border points.

\begin{definition}
	Weak Border Point (WBP)

	A point is defined to be a WBP if its nearest mean (center or One-SCOM) is not unique but less than $ K $, where $ K $ is the number of clusters.
	\label{def_border_p_weak}
\end{definition}

\begin{definition}
	Strong Border Point (SBP)
	
	A point is defined to be a SBP if its nearest mean (center or One-SCOM) is not unique and equals $ K $, where $ K $ is the number of clusters.
	\label{def_border_p_strong}
\end{definition}

Then we can process different kinds of border points with different strategies.
E.g., in Figure \ref{fig:strong_weak}a, the black dots are Strong Border Points; the black circles are Weak Border Points. In Figure \ref{fig:strong_weak}b, Weak Border Points are processed with MMJ classifier discussed in Section \ref{sec:pred}.
 
\section{Conclusion and Future Works} 
We proposed three algorithms for calculating or estimating Min-Max-Jump distance (MMJ distance), and tested several applications of it, MMJ-based K-means, MMJ-based Silhouette coefficient, and MMJ classifier. The result shows MMJ distance has good capability and potentiality in Machine Learning.  Further research may test its applications in other models, such as other clustering evaluation indices.
 
\bibliographystyle{ACM-Reference-Format}
\bibliography{mmj} 
 
\end{document}